\documentclass[final]{cvpr}

\usepackage{times}
\usepackage{epsfig}
\usepackage{graphicx}
\usepackage{amsmath}
\usepackage{amssymb}

\usepackage{multirow}
\usepackage{bm}

\usepackage[pagebackref=true,breaklinks=true,colorlinks,bookmarks=false]{hyperref}

\def\cvprPaperID{****} 
\def\confYear{CVPR 2021}

\begin{document}

\title{Representing Videos as Discriminative Sub-graphs for Action Recognition\thanks{{\small This work was performed at JD AI Research.}}}

\author{Dong Li$^{\dagger}$, Zhaofan Qiu$^{\ddagger}$, Yingwei Pan$^{\ddagger}$, Ting Yao$^{\ddagger}$, Houqiang Li$^{\dagger}$, and Tao Mei$^{\ddagger}$\\
\parbox{40em}{\small\centering $^{\dagger}$ University of Science and Technology of China, Hefei, China ~~~~~~~~~~~~$^{\ddagger}$ JD AI Research, Beijing, China}\\
{\tt\small \{dongli1995.ustc, zhaofanqiu, panyw.ustc, tingyao.ustc\}@gmail.com} \\
{\tt\small lihq@ustc.edu.cn, tmei@jd.com}
}

\maketitle
\thispagestyle{empty}

\begin{abstract}
Human actions are typically of combinatorial structures or patterns, i.e., subjects, objects, plus spatio-temporal interactions in between. Discovering such structures is therefore a rewarding way to reason about the dynamics of interactions and recognize the actions. In this paper, we introduce a new design of sub-graphs to represent and encode the discriminative patterns of each action in the videos. Specifically, we present MUlti-scale Sub-graph LEarning (MUSLE) framework that novelly builds space-time graphs and clusters the graphs into compact sub-graphs on each scale with respect to the number of nodes. Technically, MUSLE produces 3D bounding boxes, i.e., tubelets, in each video clip, as graph nodes and takes dense connectivity as graph edges between tubelets. For each action category, we execute online clustering to decompose the graph into sub-graphs on each scale through learning Gaussian Mixture Layer and select the discriminative sub-graphs as action prototypes for recognition. Extensive experiments are conducted on both Something-Something V1 \& V2 and Kinetics-400 datasets, and superior results are reported when comparing to state-of-the-art methods. More remarkably, our MUSLE achieves to-date the best reported accuracy of 65.0\% on Something-Something V2 validation set.
\end{abstract}

\section{Introduction}
The recognition of human actions is to analyze a video and identify the actions taking place in the video. In general, an action arises from interactive motion, and involves actors, objects and functional interactions in between. This characteristics motivate us to represent the structure of a video as a spatio-temporal graph. The graph nodes correspond to the volumes of actor or object regions in space-time and the edges capture the interactions between the volumes. Then, a valid question is how to reason about such graph structure to recognize the action. The difficulty originates from two aspects: 1) is it necessary to capitalize on the whole graph for reasoning? 2) considering the fact that the complexity of different actions is various, how to discover and model the discriminative patterns for each action in a unified framework?

\begin{figure}[!t]
\vspace{-0.15in}
\centering {\includegraphics[width=0.47\textwidth]{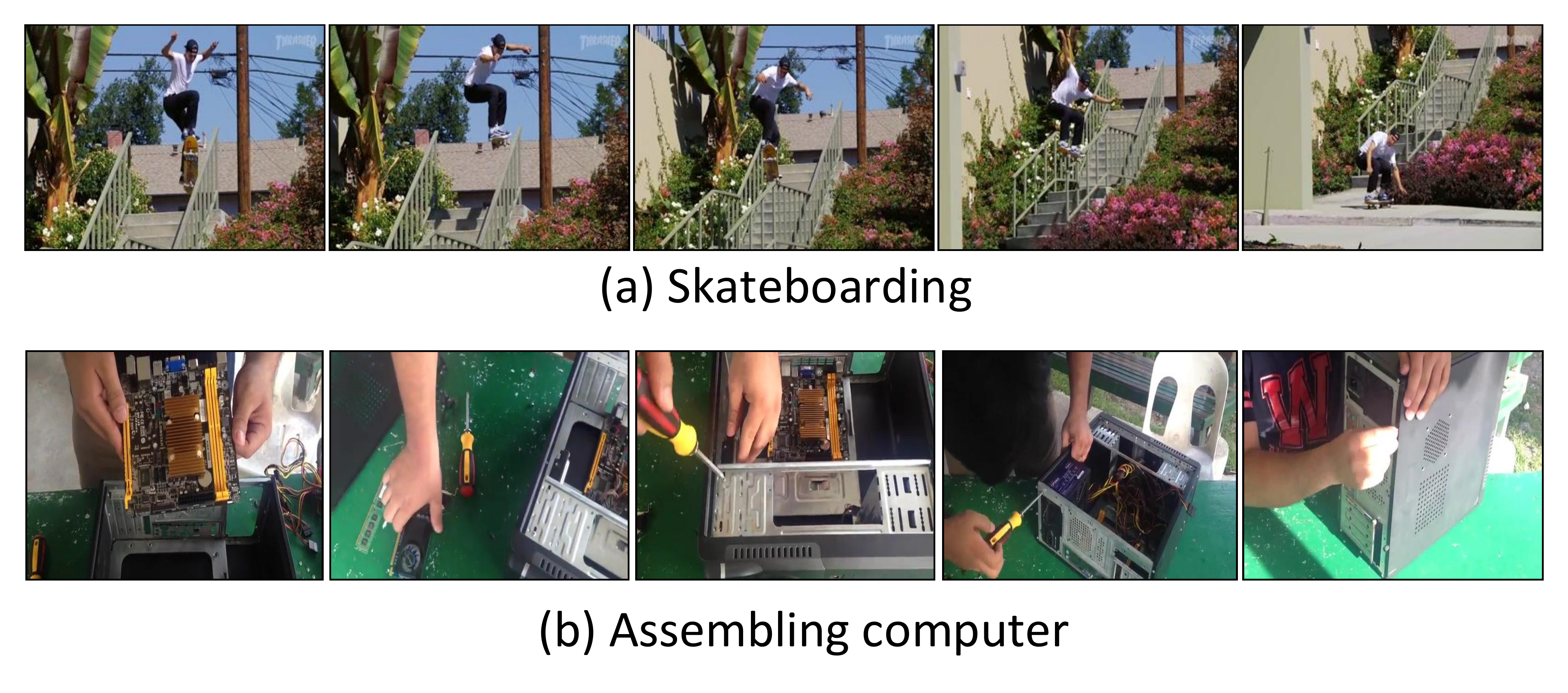}}
\vspace{-0.1in}
\caption{\small Two example videos of the action of ``skateboarding" and ``assembling computer," respectively.}
\vspace{-0.2in}
\label{fig:intro}
\end{figure}

In an effort to answer the two questions, let's look at the two action instances illustrated in Figure \ref{fig:intro}. In the upper video clip of action ``skateboarding," the objects contain ``building," ``pole," ``plants," and ``skateboard." Nevertheless, the action of ``skateboarding" only involves the actor (person), the object of ``skateboard," and the occurring interactions. In other words, the action is irrelevant to the other objects and only necessitates using a part of video structure (graph). As a result, we propose to mitigate this issue through learning discriminative sub-graphs as the prototypes of each action category rather than performing reasoning on the whole graph. The design of such prototypes takes the advantages of clustering on sub-graphs, on one hand, offers greater discriminative power in recognition, and on the other, is helpful for encoding intra-class variance. Compared to the upper video in which the action is related to only one object ``skateboard," the lower one of action ``assembling computer" correlates with more objects, including ``screwdriver," ``computer parts," and ``computer." From this view, the action ``assembling computer" is relatively more complex and thus its prototypical sub-graphs should include more nodes. To take the complexity of actions into account, the framework should enable flexibility in the scale of sub-graphs, i.e., the number of nodes in sub-graphs, to better reflect the inherent properties of different actions.

To consolidate the idea of modeling discriminative sub-graphs in videos for various actions, we present MUlti-scale Sub-graph LEarning (MUSLE) framework for action recognition. Specifically, we evenly divide an input video into a set of fixed-length video clips, which are fed into Tubelet Proposal Networks (TPN) to produce space-time actor/object tubelets. Technically, we leverage Faster R-CNN pre-trained on COCO dataset to initialize the region proposals in the first frame, and then estimate the movement of each proposal frame by frame in each clip to finally link the volume of proposals across frames as tubelets. The representation of each tubelet is the concatenation of visual features plus the coordinates of its constituent proposals. Our MUSLE takes all the tubelets from clips in a video as graph nodes and exploits dense connectivity as graph edges, which measure both semantic similarity and relative coordinate changes between every two nodes. Next, MUSLE decomposes the whole graph into multiple scales of sub-graphs. Each scale corresponds to a fixed number of nodes in the sub-graphs. We capitalize on one Gaussian Mixture Layer to interpret the distribution of all the sub-graphs from an identical action on each scale and learn $K$ Gaussian kernels. Each kernel is regarded as the discriminative sub-graph or action prototype on that scale. Note that we optimize our MUSLE framework in an end-to-end manner. During inference, we compute the similarity between sub-graphs extracted from the test video and action prototypes across all the scales and actions, and take the class of action prototype with the highest similarity as prediction.

The main contribution of this work is the proposal of representing video structure as a space-time graph and eventually discovering the discriminative sub-graphs for action recognition. This also leads to the elegant views of how to perform end-to-end learning of the discriminative sub-graphs, and how to nicely present the complexity of different actions in the reasoning process, which are problems not yet fully understood. We demonstrate the effectiveness of our design, i.e., MUSLE framework, on Something-Something V1\&V2 and Kinetics-400 datasets, and superior performances are reported in the experiments.

\section{Related Works}
\textbf{Action Recognition} is a fundamental computer vision task and has been extensively studied recently. Early approaches usually rely on hand-crafted features, which detect spatio-temporal interest points and then describe these points with local representations \cite{wang2011action,wang2013action}. With the tremendous success of deep convolution networks on image-based classification tasks \cite{he2016deep,russakovsky2015imagenet,simonyan2015very,szegedy2015going}, researchers started to explore the application of deep networks on video action recognition task \cite{fan2019more,kwon2020motionsqueeze,long2019gaussian,long2020learning,yao2021seco}. In \cite{simonyan2014two}, the famous two-stream architecture is devised by applying two 2D CNN architectures separately on visual frames and staked optical flows. This two-stream architecture is further extended by exploiting sparse temporal sampling \cite{wang2016temporal}, convolutional fusion \cite{feichtenhofer2016convolutional}, convolutional encoding \cite{diba2017deep,qiu2017deep}, and spatio-temporal attention \cite{li2018unified}. \cite{yue2015beyond} highlights a drawback of two-stream architecture that exploits a standard image CNN instead of a specialized network for training videos, which makes the two-stream network unable to capture long-term temporal information. To address this issue, Tran \emph{et al.} \cite{tran2015learning} propose a 3D CNN (i.e., C3D) for learning video representation, which performs 3D convolutions on adjacent frames to jointly model the spatial and temporal features. Compared to 2D CNN, C3D holds much more parameters and is difficult to obtain good convergence. Consequently, I3D \cite{carreira2017quo} further takes advantage of ImageNet pretraining by inflating 2D CNN into 3D. To reduce the heavy computations of 3D CNNs, several methods are proposed to find the trade-off between precision and speed \cite{lin2019tsm,qiu2017learning,tran2018closer,xie2018rethinking,zolfaghari2018eco}. For example, P3D \cite{qiu2017learning} and R(2+1)D \cite{tran2018closer} decompose the 3D convolution into a 2D spatial convolution and a 1D temporal convolution. TSM \cite{lin2019tsm} shifts the features across the channel dimension to perform temporal modeling. LGD \cite{qiu2019learning} further devises a two-pathway architecture to learn local and global representations in parallel. In this work, we imitate the design of LGD in the backbone of tubelet feature extractor, but our measure of interactions among actors or objects could be readily integrated into any advanced networks.

\textbf{Graphical Models} have been proven to be helpful for relation reasoning in various computer vision tasks, such as semantic segmentation \cite{bertasius2017convolutional,chandra2017dense,liang2018symbolic} and image captioning \cite{yang2019auto,yao2018exploring,zhong2020comprehensive}. Neural networks that operate on graphs have previously been introduced as a form of RNN in early works \cite{deng2016structure,li2015gated,scarselli2008graph}. Thanks to the proposal of Graph Convolutional Networks (GCN) \cite{kipf2017semi} which generalize the convolutional operation to deal with graph-structured data, many graph-based methods have been devised for action recognition in recent years, especially for skeleton-based action recognition. For example, ST-GCN \cite{yan2018spatial} proposes a spatio-temporal graph to model the structured information among the human joints. Different from skeleton-based action recognition where the skeleton data can be naturally seen as graph structure, generic action recognition methods employ GCNs to model the relations between fixed regions or objects. For example, \cite{chen2019graph,li2018beyond,wang2020region} adopts GCN to build a reasoning module to model the relations between disjoint and distant regions. \cite{li2019long,wang2018videos} takes dense object proposals as graph nodes and learns the relations between them. \cite{li2020adaptive} treats each object proposal detected in the sample frames as a graph node and then searches adaptive network structures to model the object interactions. Unlike aforementioned graphical techniques for action recognition which perform relation reasoning over the complete graph, our work contributes by discovering the discriminative sub-graphs across different scales for facilitating spatio-temporal reasoning.

\begin{figure*}[!tb]
\centering {\includegraphics[width=0.95\textwidth]{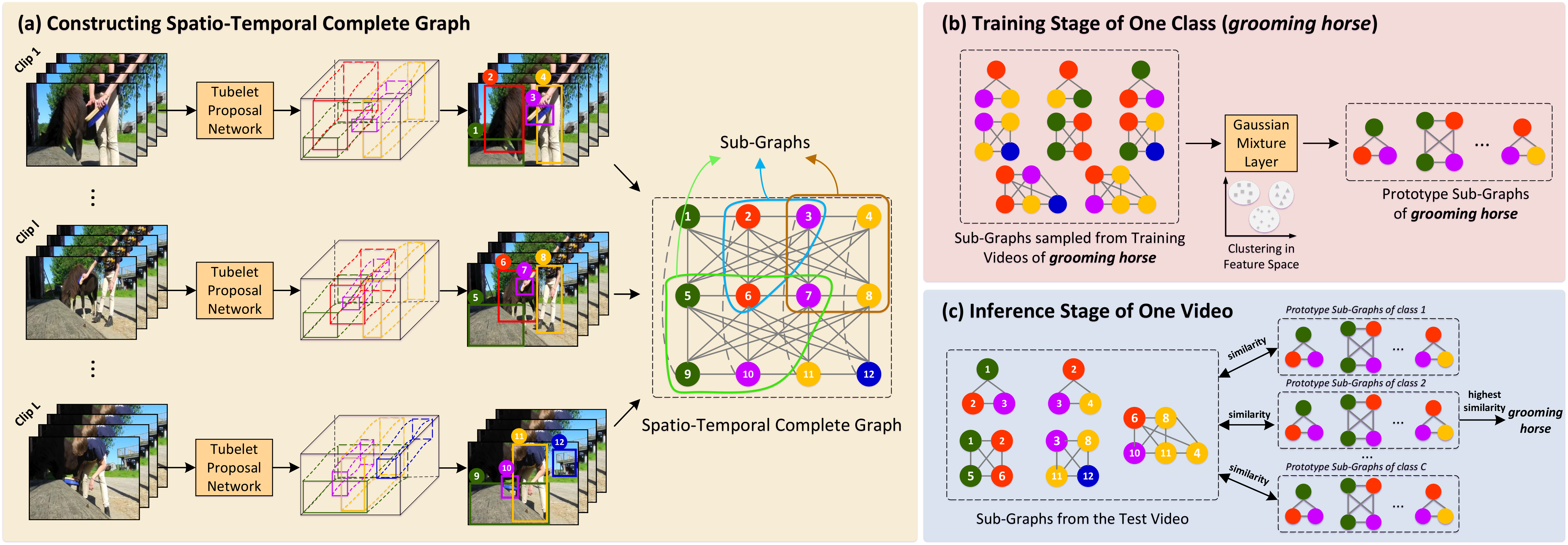}}
\caption{\small An overview of our MUSLE framework. (a) Given an input video, $L$ video clips is produced by evenly dividing this video. All clips are fed into Tubelet Proposal Networks to produce space-time actor/object tubelets in parallel. Next, by taking all tubelets as graph nodes and exploiting dense connectivity as edges, we build a spatio-temporal complete graph to perform reasoning. After that, MUSLE decomposes the whole graph into sub-graphs with multiple scales. (b) During training, conditioned on all the sub-graphs from an identical action on each scale, a Gaussian Mixture Layer is learnt to discover the discriminative sub-graphs or action prototypes for that action. (c) At inference, after extracting all sub-graphs across different scales from the test video, we calculate the similarity between all sub-graphs and action prototypes across all the scales and actions. The class of action prototype with the highest similarity is thus taken as the prediction.}
\label{fig:framework}
\vspace{-0.15in}
\end{figure*}

\section{Multi-scale Sub-graph Learning}
In this paper, we devise a MUlti-scale Sub-graph LEarning (MUSLE) framework to exploit the discriminative sub-graphs across different scales for facilitating spatio-temporal reasoning. Figure \ref{fig:framework} depicts an overview of our architecture for action recognition. The training process of MUSLE consists of three main components: tubelet generation, spatio-temporal complete graph construction, and discriminative sub-graph extraction. Specifically, given the set of video clips evenly divided from the input video, we leverage Tubelet Proposal Networks (TPN) to generate space-time actor/object tubelets for each video clip. After that, a spatio-temporal complete graph is constructed by taking all the tubelets within a video as graph nodes and exploiting dense connectivity as
edges. We further decompose such complete graph into a series of sub-graphs across different scales. For each scale, a Gaussian Mixture Layer is learnt to extract the discriminative sub-graphs or action prototypes. During inference, we extract sub-graphs from each test video, and further measure the similarity between the extracted sub-graphs and action prototypes across all the scales and actions. The action category of action prototype with the highest similarity is finally taken as the prediction.

\subsection{Tubelet Proposal Networks}
The Tubelet Proposal Networks (TPN) targets for producing the actor/object tubelets, i.e., the volume of actor/object proposals across consecutive frames, within each video clip. Here we remould TPN in \cite{li2018recurrent} by migrating it from actor-only tubelet generation to our scenario of both actor and object tubelet generation. Concretely, we first capitalize on an actor/object detector to find actor/object proposals in the start frame of each clip. The TPN further estimates the movements of each proposal in current frame to produce the corresponding proposal in the subsequent frame. The volume of proposals across consecutive frames are finally linked as the tubelets within that clip.

Formally, given a video clip with $T$ frames, we firstly adopt Faster R-CNN pre-trained on COCO dataset \cite{lin2014microsoft} to detect the actor/object region proposals in the first frame. We keep the top-$M$ proposals $B_1=\{b_1^m|m=1,...,M\}$  with the highest detection scores as the proposal set of the first frame. Next, conditioned on the $t$-th frame $I_t$ and its proposal set $B_t$, TPN produces the proposal set $B_{t+1}$ for the next frame $I_{t+1}$ recurrently. Specifically, for each proposal $b_t^m$ in $t$-th frame $I_t$, we estimate the movement of $b_t^m$ in the next frame $I_{t+1}$ depending on the visual features of the same location of $b_t^m$ across the two frames. Here we directly define the outputs of RoI pooling in $I_t$ and $I_{t+1}$ for the same location of $b_t^m$ as the visual features $\mathbf{F}_t^m$ and $\mathbf{F}_{t+1}^m\in \mathbb{R}^{w\times h\times d}$, where $w$, $h$ and $d$ are width, height and channel numbers. A bilinear kernel is further utilized to capture the pairwise correlations and model spatial interactions between $\mathbf{F}_t^m$ and $\mathbf{F}_{t+1}^m$ for movement estimation:
\begin{align}
\begin{split}
\label{eq:bilinear}
\langle \mathbf{F}_t^m, \mathbf{F}_{t+1}^m \rangle_2 &= \frac{1}{S^2} \sum_{i=1}^{S}\sum_{j=1}^{S} \left \langle \mathbf{F}_{t,i}^m, \mathbf{F}_{t+1,j}^m \right \rangle_2 \\
                                                     &\approx \frac{1}{S^2} \sum_{i=1}^{S}\sum_{j=1}^{S} \left \langle \Phi (\mathbf{F}_{t,i}^m),  \Phi (\mathbf{F}_{t+1,j}^m) \right \rangle,
\end{split}
\end{align}
where $S=w\times h$ is the number of spatial locations, and $\left \langle \cdot , \cdot \right \rangle_2$ is the bilinear kernel. We approximate bilinear kernel via Tensor Sketch Projection $\Phi$ \cite{gao2016compact}, aiming to reduce the feature space dimension. The kernelized representation is then fed into a regression layer to predict the movement of $b_t^m$, leading to the corresponding proposal $b_{t+1}^m$ in $I_{t+1}$. Accordingly, a set of proposals in the subsequent frames are obtained by estimating the movement of each proposal frame by frame, which are linked as the output tubelet.

\subsection{Spatio-Temporal Complete Graph}

With all the generated tubelets of each input video via TPN, we next present how to construct a spatio-temporal complete graph, that fully unfolds the inherent spatial and temporal interactions among all tubelets for reasoning. Specifically, we are given the set of $L$ video clips and each video clip contains $M$ tubelets, leading to the tubelet set $\mathcal{V}=\{v_i\}^{L \times M}_{i=1}$. By treating each tubelet as one graph node, we can build the spatio-temporal complete graph $\mathcal{G}=(\mathcal{V}, \mathcal{E})$, where $\mathcal{E}$ denotes the set of densely connected graph edges between every two nodes.
As depicted in Figure \ref{fig:framework}, we number all tubelets according to their spatio-temporal positions in the video: the spatially closer to the top left and the temporally earlier the tubelet, the smaller the number of that tubelet.

\textbf{Graph Node.} For each node $v \in \mathcal{V}$, we collect both visual \& spatial information and represent it as the concatenation of tubelet visual feature $\mathbf{f}_v^{visual}$ and the constituent bounding box coordinates $\mathbf{f}_v^{coord}$: $\mathbf{f}_v=[\mathbf{f}_v^{visual}, \mathbf{f}_v^{coord}]$.

To extract the visual feature of each tubelet in $l$-th clip, we first feed this clip ($T$ consecutive frames) into a standard 3D CNN. The output of 3D CNN is a feature map with the size of $T \times W\times H\times D$ ($T$: temporal dimension; $H \times W$: spatial dimensions; $D$: channel number). Note that we do not utilize any temporal downsampling layers (either temporal pooling or time-strided convolutions) in 3D CNN, preserving the temporal fidelity as much as possible. Next, conditioned on the spatial coordinates of the bounding boxes across frames in tubelet, we perform RoI pooling over the output feature map of 3D CNN, followed by global max pooling to produce a $D$-dimensional visual representation $\mathbf{f}_v^{visual}$ of that tubelet.

Considering that some actions are naturally rooted in the geometric properties (e.g., 2D location) of actors/objects, we further exploit the coordinates of constituent bounding boxes in the tubelet to represent each node. Specifically, the 2D location of the bounding box in each frame can be denoted as the 4-dimensional feature vector consisting of its center coordinates plus the height and width values. Thus, we represent the holistic location feature of each tubelet as a $4T$-dimensional feature vector $\widehat{\mathbf{f}}_{v}^{coord}$, i.e., the concatenation of 2D locations across all frames. The holistic location feature of each tubelet is finally transformed into $\mathbf{f}_v^{coord}$ via a Multi-Layer Perception (MLP).

\textbf{Graph Edge.} Most existing works (e.g., \cite{wang2018videos}) directly encapsulate the visual similarity and geometrical overlap between two nodes into a scalar weight to represent each edge. This way apparently leaves the relative spatial relation or distance in between under-explored, and might result in failure of reasoning for distance-sensitive actions. Therefore, for each edge $e \in \mathcal{E}$, we derive a particular form of edge representation $\mathbf{f}_e=[f_e^{sem},\mathbf{f}_e^{coord}]$, which concatenates the semantic similarity $f_e^{sem}$ and relative coordinate changes $\mathbf{f}_e^{coord}$ between nodes. The rational behind is to encourage the edge feature $\mathbf{f}_e$ to reflect both semantic and relative spatial relations between tubelets, and thus facilitate the reasoning process.

Concretely, given two nodes $v_i$ and $v_j (i<j)$, we measure the semantic similarity between them as
\begin{equation}\label{Eq:edge}
f_{e_{ij}}^{sem}={\varphi(\mathbf{f}_{v_i}^{visual})}^\top\varphi(\mathbf{f}_{v_j}^{visual}),
\end{equation}
where $\mathbf{f}_{v_i}^{visual}$ and $\mathbf{f}_{v_j}^{visual}$ are the visual features of $v_i$ and $v_j$, respectively. $\varphi(\cdot)$ denotes the feature transformation function.
Moreover, depending on the holistic location features of $v_i$ and $v_j$ ($\widehat{\mathbf{f}}_{v_i}^{coord}$, $\widehat{\mathbf{f}}_{v_j}^{coord}$), we directly calculate the relative coordinate changes as $\widehat{\mathbf{f}}_{e_{ij}}^{coord}=\widehat{\mathbf{f}}_{v_i}^{coord}-\widehat{\mathbf{f}}_{v_j}^{coord}$. We further employ a MLP over $\widehat{\mathbf{f}}_{e_{ij}}^{coord}$ to obtain the final representation of relative coordinate changes $\mathbf{f}_{e_{ij}}^{coord}$.

\subsection{Discriminative Sub-graphs}
One natural way to perform spatio-temporal reasoning across actors/objects is to model the interactions between tubelets over the space-time complete graph. However, such complete graph might result in some irrelevant objects with regard to different action categories, and the overall stability of spatio-temporal reasoning will be inevitably affected. To alleviate this issue, our MUSLE learns to decompose the complete graph into a series of discriminative sub-graphs across different scales. The rational behind is to interpret each discriminative sub-graph as one prototype of each action category, that conveys greater discriminative power for the recognition of that action. Technically, we first extract multi-scale sub-graphs from the complete graph. Here each scale corresponds to a fixed number of nodes (i.e., 3, 4, or 5 nodes) in the sub-graphs. After that, for all sub-graphs from an identical action under each scale, we leverage a Gaussian Mixture Layer to interpret the distribution of them via Gaussian Mixture Model (GMM). $K$ Gaussian kernels are thus achieved to represent the discriminative sub-graphs/action prototypes on that scale.

\textbf{Gaussian Mixture Layer.} Inspired by deep autoencoding Gaussian Mixture Model \cite{fan2020video,zong2018deep}, we utilize a Gaussian Mixture Layer to perform online clustering over all the sub-graphs to learn Gaussian kernels, rather than an offline training of GMM through standard EM algorithm \cite{huber2004robust}. Formally, given a set of $N$ sub-graphs from an identical action on each scale, we represent each sub-graph as a feature vector \textbf{x} by concatenating all the node and edge representations within that sub-graph. Note that all nodes/edges are concatenated according to their assigned numbers (from small to large). Next, with the inputs of each sub-graph representation \textbf{x} and an integer $K$ (the number of mixture components/Gaussian kernels), we first leverage a MLP to predict the mixture membership of \textbf{x}:
\begin{equation}
\bm{\alpha} = MLP(\mathbf{x};\bm{\theta}),~~~~~~\hat{\bm{\gamma}} = Softmax(\bm{\alpha}),
\end{equation}
where $\bm{\theta}$ is the network parameter and $\hat{\bm{\gamma}}$ is a $K$-dimensional vector for the soft mixture membership prediction. After that, conditioned on all the mixture membership predictions of the $N$ sub-graphs, the parameters of the $k^{th}$ ($1 \leq k \leq K$) Gaussian kernel in GMM can be estimated as:
\begin{align}
\begin{split}
\hat{\phi}_k &= \sum_{n=1}^{N}\frac{\hat{\gamma_{nk}}}{N},~~~~~~~~\hat{\bm{\mu}}_k = \frac{\sum_{n=1}^{N}\hat{\gamma}_{nk}\mathbf{x}_n}{\sum_{n=1}^{N}\hat{\gamma}_{nk}} \\
\hat{\bm{\Sigma}}_k                &= \frac{\sum_{n=1}^{N}\hat{\gamma}_{nk}(\mathbf{x}_n-\hat{\bm{\mu}}_k)(\mathbf{x}_n-\hat{\bm{\mu}}_k)^\top}{\sum_{n=1}^{N}\hat{\gamma}_{nk}}
\end{split}
\end{align}
where $\hat{\bm{\gamma}}_n$ is the mixture membership prediction for the sub-graph $\mathbf{x}_n$. $\hat{\bm{\phi}}_k$, $\hat{\bm{\mu}}_k$, $\hat{\bm{\Sigma}}_k$ denotes the mixture weight, mean, and covariance matrix for the $k^{th}$ Gaussian kernel in GMM, respectively. Accordingly, with the estimated parameters of all Gaussian kernels, the possibility of sample $\mathbf{x}$ belonging to the action category can be calculated as:
\begin{equation} \label{Eq:cla}
p(\mathbf{x}) = \sum_{k=1}^{K}\hat{\phi}_k\frac{exp(-\frac{1}{2}(\mathbf{x}-\hat{\bm{\mu}}_k)^\top\hat{\bm{\Sigma}}_k^{-1}(\mathbf{x}-\hat{\bm{\mu}}_k))}{\sqrt{|2\pi\hat{\bm{\Sigma}}_k|}},
\end{equation}
where $|\cdot|$ denotes the determinant of a matrix.

\textbf{Training Objective.} At the training stage, we formulate the overall objective over all the $N$ sub-graphs on each scale in a batch~as:
\begin{equation}
\mathcal{L}(\bm{\theta})= -\frac{1}{N}\sum_{n=1}^{N}log~p(\mathbf{x}_n) + \lambda R(\hat{\bf{\Sigma}}),
\end{equation}
where $p(\mathbf{x}_n)$ is the predicted possibility of each sub-graph $\mathbf{x}_n$ calculated as in Eq. \ref{Eq:cla}. $R(\hat{\bf{\Sigma}}) = \sum_{k=1}^{K}\sum_{i=1}^{D_{\mathbf{x}}}\frac{1}{\hat{\Sigma}_{kii}}$ ($D_{\mathbf{x}}$: feature dimension of $\mathbf{x}$) is a regularization of covariance matrices $\hat{\bf{\Sigma}}$, which restricts the diagonal entries in $\hat{\bf{\Sigma}}$ converge to reasonable solution rather than 0. $\lambda$ denotes the parameter for balancing the two parts, and we set it as 0.05 in practice. Since every operation within the Gaussian Mixture Layer is differentiable, we can easily propagate gradient from the Gaussian Mixture Layer to the tubelet feature extractor, and thus the two modules can be jointly optimized in an end-to-end manner.

Please note that instead of fixing the number of Gaussian kernels $K$ under each scale for each action category, we employ a dynamic strategy to automatically update $K$ along training process. Specifically, we initialize $K$ with a large value (6 in our case), and $K$ will be decreased until all mixture weights of Gaussian kernels are higher than threshold $th$ ($th=0.02$) after one training epoch.

\textbf{Inference.} During inference, for each testing video, we first build the corresponding spatio-temporal complete graph, which is further decomposed into a group of sub-graphs across all the three scales. Next, for each action category, we utilize Eq. \ref{Eq:cla} to estimate the probability of each sub-graph belonging to that category conditioned on the learnt action prototypes. The highest probability across all sub-graphs under three scales is thus taken as the classification score of the testing video for this category. Finally, the action category with the highest classification score is regarded as the predicted category of the testing video.

\section{Experiments}

\begin{table*}[tb] \footnotesize
\centering
\caption{\small Performance contribution of each component in MUSLE. Experiments are conducted on Something-Something V2 validation set.}
\label{table:ablation}
\begin{tabular}{l|ccccc|cl}
\hline
\textbf{Method}                                  & \textbf{Graph} & \textbf{Sub-graph} & \textbf{Node Coord} & \textbf{Edge Coord} & \textbf{Multi-scale} & \textbf{Top-1} & \textbf{Top-5} \\ \hline \hline
Base                                             &                &                    &                     &                     &                      & 60.8           & 86.5           \\ \hline
Base+GCN \cite{wang2018videos}                   & $\surd$        &                    &                     &                     &                      & 61.3           & 87.1           \\
Base+GloRe \cite{chen2019graph}                  & $\surd$        &                    &                     &                     &                      & 61.9           & 87.3           \\ \hline
Base                                             &                &                    &                     &                     &                      &                &                \\
\quad \quad \quad \quad \quad (3 nodes)          & $\surd$        &  $\surd$           &                     &                     &                      & 62.7           & 88.0           \\
+Sub-graph (4 nodes)                             & $\surd$        &  $\surd$           &                     &                     &                      & 63.2           & 88.3           \\
\quad \quad \quad \quad \quad (5 nodes)          & $\surd$        &  $\surd$           &                     &                     &                      & 62.2           & 87.5               \\
+Node Coord (4 nodes)                            & $\surd$        &  $\surd$           &  $\surd$            &                     &                      & 63.5           & 88.7               \\
+Edge Coord (4 nodes)                            & $\surd$        &  $\surd$           &  $\surd$            &  $\surd$            &                      & 64.1           & 89.3               \\ \hline
\textbf{MUSLE}                                   & $\surd$        &  $\surd$           &  $\surd$            &  $\surd$            &  $\surd$             & \textbf{65.0}  & \textbf{90.1}         \\ \hline
\end{tabular}
\vspace{-0.15in}
\end{table*}

In this section, we empirically evaluate our MUSLE on two challenging action recognition benchmarks: Something-Something V1\&V2 \cite{goyal2017something} and Kinetics-400 \cite{carreira2017quo}.
\subsection{Datasets}

\textbf{Something-Something} is one of the largest video datasets that focus on human-object interaction scenarios. The dataset contains 174 fine-grained categories of common human-object interactions with diverse objects and viewpoints. The recognition of them is challenging, which requires fine-grained understanding of the activity to distinguish similar actions within a group, e.g., ``pushing something so that it falls off the table'' and ``pushing something so that it almost falls off but doesn't.'' The first version (V1) consists of around 108k videos in total, including 86k for training, 11k for validation, and 11k for testing. The second version (V2) further increases the video number to 220k. The average video length is 4.0 seconds and all videos are captured from object-centric view with fairly clean backgrounds. Thus, few scene contexts is required to be exploited for action recognition in this dataset, making it a~natural choice of benchmark for evaluating spatio-temporal reasoning in videos (e.g., \cite{li2020adaptive,wang2018videos}). Similarly, we mainly report the results of our MUSLE on this dataset to evaluate the capacity of spatio-temporal reasoning.

\textbf{Kinetics-400} is a standard large-scale benchmark for action recognition, covering 400 action classes. It contains around 246k training videos and 20k validation videos. Each video in this dataset is 10-second short clip trimmed from the raw YouTube video. Here we additionally involve this dataset to further demonstrate the generalization of our proposal for action recognition.

\subsection{Implementation Details}

\textbf{Training.} At training stage, we adopt the same strategy as in TSN \cite{wang2016temporal} to train our MUSLE framework. Specifically, given an input video, we first divide it into $L$ ($L=4$ in our case) segments with equal durations, aiming to perform long-range temporal structure modeling. Then, for each segment, we randomly sample one clip consisting of $16$ consecutive frames. The size of the short side of sampled frames is fixed to $256$. We augment the data during training by scale and aspect-ratio jittering. For tubelet generation process, we utilize Faster R-CNN \cite{ren2015faster} pre-trained on COCO dataset \cite{lin2014microsoft} to detect objects in the first frame of each clip. Meanwhile, we pre-train the Tubelet Proposal Network \cite{li2018recurrent} on ImageNet VID dataset \cite{russakovsky2015imagenet} to estimate the movement of each object region frame by frame in each clip and fix its parameters during training. We select the top 8 tubelets with the highest detection scores in each clip to build the spatio-temporal complete graph, leading to 32 nodes in the complete graph for the whole video (consisting of 4 sampled clips). We capitalize on LGD network \cite{qiu2019learning} as the tubelet feature extractor, which represents each tubelet as the outputs from the last convolutional layer with RoI Pooling. The parameters of the LGD network is initialized with the ImageNet pre-trained ResNet-50 model.

We implement MUSLE mainly on Caffe \cite{jia2014caffe}. The whole framework is trained on four Tesla P40 GPUs via Stochastic Gradient Descent (SGD) with a mini-batch of 16 videos. The momentum and weight decay are set to 0.9 and 0.0005, respectively. We set the initial learning rate as 0.01, which is further divided by 10 after every 20 epochs. The training is stopped after 50 epoches.

\textbf{Inference}. During inference, we evenly sample 4 clips from each test video. For each frame in the clips, we follow the strategy in \cite{jiang2019stm,li2020tea} to resize the shorter size as 256 by maintaining the aspect ratio. We utilize the learnt Gaussian Mixture Model to compute the similarity between the sub-graphs extracted from the test video and action prototypes across all the scales and actions. The class of action prototype with the highest similarity is taken as the prediction.

\subsection{Ablation Study of MUSLE}

\begin{table*}[tb] \footnotesize
\centering
\caption{\small Comparison results of MUSLE with other state-of-the-art methods on Something-Something V1 \& V2.}
\label{table:sth}
\begin{tabular}{l|c|c|c|c|c|c|ccc}
\hline
\multirow{2}{*}{\textbf{Method}} & \multirow{2}{*}{\textbf{Backbone}} & \multirow{2}{*}{\textbf{Pre-train}} & \multirow{2}{*}{\textbf{\#Frames}} & \multicolumn{3}{c|}{\textbf{Something-Something V1}} & \multicolumn{3}{c}{\textbf{Something-Something V2}}                          \\ \cline{5-10}
                                 &                                    &                                     &                                   & top-1 val       & top-5 val       & top-1 test       & \multicolumn{1}{c|}{top-1 val} & \multicolumn{1}{c|}{top-5 val} & top-1 test \\ \hline \hline
TSN \cite{wang2016temporal}      & BNInception                        & ImageNet                            & 16                                & 19.7            & 46.6            & -                & \multicolumn{1}{c|}{27.8}      & \multicolumn{1}{c|}{57.6}      & -          \\
TRN \cite{zhou2018temporal}      & BNInception                        & ImageNet                            & 8                                 & 34.4            & -               & 33.6             & \multicolumn{1}{c|}{48.8}      & \multicolumn{1}{c|}{77.6}      & 50.9       \\
DualAtt \cite{xiao2019reasoning} & BNInception                        & ImageNet                            & 8                                 & -               & -               & -                & \multicolumn{1}{c|}{51.6}      & \multicolumn{1}{c|}{80.3}      & 54.0       \\
TRG \cite{zhang2020temporal}     & BNInception                        & ImageNet                            & 16                                & 45.9            & 74.9            & -                & \multicolumn{1}{c|}{56.7}      & \multicolumn{1}{c|}{79.9}      & -       \\ \hline \hline
I3D \cite{carreira2017quo}       & ResNet-50                          & Kinetics                            & 32                                & 41.6            & 72.2            & -                & \multicolumn{1}{c|}{-}         & \multicolumn{1}{c|}{-}         & -          \\
I3D+GCN \cite{wang2018videos}    & ResNet-50                          & Kinetics                            & 32                                & 43.3            & 75.1            & -                & \multicolumn{1}{c|}{-}         & \multicolumn{1}{c|}{-}         & -          \\
NL I3D+GCN \cite{wang2018videos} & ResNet-50                          & Kinetics                            & 32                                & 46.1            & 76.8            & 45.0             & \multicolumn{1}{c|}{-}         & \multicolumn{1}{c|}{-}         & -          \\
S3D \cite{xie2018rethinking}     & Inception                          & ImageNet                            & 64                                & 48.2            & 78.7            & 42.0             & \multicolumn{1}{c|}{-}         & \multicolumn{1}{c|}{-}         & -          \\
ECO \cite{zolfaghari2018eco}     & BNInc+ResNet-18                    & Kinetics                            & 16                                & 41.4            & -               & -                & \multicolumn{1}{c|}{-}         & \multicolumn{1}{c|}{-}         & -          \\
TSM \cite{lin2019tsm}            & ResNet-50                          & Kinetics                            & 16                                & 47.2            & 77.1            & 46.0             & \multicolumn{1}{c|}{63.4}      & \multicolumn{1}{c|}{88.5}      & 64.3       \\
STM \cite{jiang2019stm}          & ResNet-50                          & ImageNet                            & 16                                & 50.7            & 80.4            & 43.1             & \multicolumn{1}{c|}{64.2}      & \multicolumn{1}{c|}{89.8}      & 63.5       \\
TEA \cite{li2020tea}             & ResNet-50                          & ImageNet                            & 16                                & 51.9            & 80.3            & -                & \multicolumn{1}{c|}{-}         & \multicolumn{1}{c|}{-}         & -          \\
GSM \cite{sudhakaran2020gate}    & InceptionV3                        & ImageNet                            & 16                                & 50.6            & -               & -                & \multicolumn{1}{c|}{-}         & \multicolumn{1}{c|}{-}         & -          \\
ASS \cite{li2020adaptive}        & ResNet-50                          & ImageNet                            & 32                                & 51.4            & -               & -                & \multicolumn{1}{c|}{63.5}      & \multicolumn{1}{c|}{-}         & -          \\ \hline
MUSLE                            & ResNet-50                          & ImageNet                            & 16                                & \textbf{52.5}   & \textbf{81.6}   & \textbf{47.4}    & \multicolumn{1}{c|}{\textbf{65.0}}      & \multicolumn{1}{c|}{\textbf{90.1}}      & \textbf{65.0}         \\ \hline
\end{tabular}
\vspace{-0.15in}
\end{table*}

Here, we investigate how each design in MUSLE influences the overall performance. We start from a basic model (named \textbf{Base}) by feeding each video clip into LGD backbone and simply taking the output score as clip-level score. The averaged score over all clips is thus exploited for action classification. Base model solely capitalizes on the clip-level holistic features, while leaving the object-level relations unexploited. As an alternative, \textbf{GCN} \cite{wang2018videos} leverages Tubelet Proposal Networks (TPN) to produce space-time actor/object tubelets for each clip. The representation of each tubelet is extracted through LGD backbone. After that, all the tubelets from clips within a video are taken as the graph nodes, which are connected with similarity relations in between. Graph Convolutional Network (GCN) is further utilized to perform spatio-temporal reasoning over the whole graph. The final video-level representation is thus obtained by performing mean pooling over all the refined graph nodes. \textbf{GloRe} \cite{chen2019graph} upgrades GCN with global reasoning unit, which conducts highly efficient global reasoning over the whole graph by projecting aggregated features over coordinate space into an interaction space. Instead of reasoning on the whole graph (e.g., GCN and GloRe), the run of \textbf{Base+Sub-graph} decomposes the whole graph (with only semantic similarity relations) into single-scale sub-graphs. For one specific scale, we utilize Gaussian Mixture Layer to learn the discriminative sub-graphs/action prototypes for action recognition. Please note that in Base+Sub-graph, we directly take the visual feature of LGD backbone to represent each tubelet. \textbf{Node Coord} further enhances tubelet representation by concatenating the visual feature and the coordinates of constituent proposals. In \textbf{Edge Coord}, the graph edges are strengthened to measure not only the semantic similarity between every two nodes, but also the relative coordinate changes in between. \textbf{Multi-scale} unifies the discriminative sub-graphs across all the three scales for action recognition.

Table \ref{table:ablation} details the performances across different ways of reasoning about graph structure for action recognition on Something-Something V2.
Specifically, for Base model, the use of original clip-level holistic features in general achieves a good performance. As expected, by additionally modeling the object-level relations over the whole spatio-temporal graph for recognizing actions, Base+GCN and Base+GloRe exhibit better performances than Base model. This generally verifies the merit of reasoning about spatio-temporal graph structure within videos. Nevertheless, performing reasoning over the whole graph would inevitably result in more irrelevant objects, which may affect the overall stability of reasoning process. To address this issue, our unique design of Sub-graph enables the learning of single-scale discriminative sub-graphs as the prototypes of each action category for recognition. Here we vary the scale (i.e., node number) of sub-graphs in the range of $\{3,4,5\}$, and the results of Base+Sub-graph under different scales consistently outperform Base+GCN and Base+GloRe across all metrics. The best performances (Top-1: 63.2\%, Top-5: 88.3\%) are attained when the node number of each sub-graph is set to 4. The results clearly highlight the advantage of exploiting sub-graphs in videos to represent and encode the discriminative patterns of each action. Furthermore, Node Coord and Edge Coord, which integrates Sub-graph with spatial-aware cues (i.e., information about proposal coordinates or relative coordinate changes), contributes an Top-1 accuracy increase of 0.3\% and 0.6\%, respectively. This demonstrates that Node Coord and Edge Coord are very practical choices to enhance the capacity of spatio-temporal reasoning, especially for the recognition of distance-sensitive actions (e.g., ``moving something and something \texttt{away} from each other'' versus ``moving something and something \texttt{closer} to each other'').
The integration of all sub-graphs across three scales, i.e., our MUSLE, reaches the highest performances for action recognition. The performance boosts basically indicate the advantage of multi-scale aggregation of sub-graphs during reasoning, that enables flexibility in the scale of sub-graphs to better reflect the inherent properties of various actions.

\subsection{Comparisons with State-of-the-Arts}

\textbf{Something-Something.}
Table \ref{table:sth} summarizes the quantitative results of our MUSLE on both Something-Something V1 and V2 datasets. We compare MUSLE with several existing state-of-the-art action recognition methods, which can be grouped into two directions: 2D CNNs based techniques (e.g., TSN, TRN, and TRG) and 3D/(2+1)D CNNs based approaches (e.g., I3D, I3D+GCN, S3D, and ASS). Overall, the results under the same
backbone (ResNet-50) across different datasets consistently demonstrate that our MUSLE exhibits better performances than other 2D and 3D/(2+1)D CNNs based models. In particular, the Top-1 val accuracy of MUSLE can achieve 52.5\% and 65.0\% on Something-Something V1 and V2 respectively, which makes 0.6\% and 0.8\% absolute improvements over the best competitors TEA and STM. By modeling temporal orders across frames for reasoning, TRN achieves better performances than TSN which simply averages the features of temporal frames. Moreover, DualAtt improves TRN by additionally capturing human-object interactions via attention mechanism. TRG further leads to a performance boost by constructing a temporal relation graph to capture the long-range temporal dependencies across frames. For 3D/(2+1)D CNNs based approaches, I3D+GCN by exploiting the relations between objects for spatio-temporal reasoning over the whole graph, outperforms I3D. Different from the manually-designed graph-based reasoning module in I3D+GCN and NL I3D+GCN, ASS automatically searches adaptive interaction modeling structures for reasoning on space-time graph, and achieves better performance. Nevertheless, the performances of ASS are still lower than our MUSLE, which uniquely discovers the discriminative sub-graphs for action recognition. This confirms the advantage of decomposing the whole graph into sub-graphs and modeling the discriminative ones for spatio-temporal reasoning in MUSLE.

\begin{figure}[!t]
\centering {\includegraphics[width=0.45\textwidth]{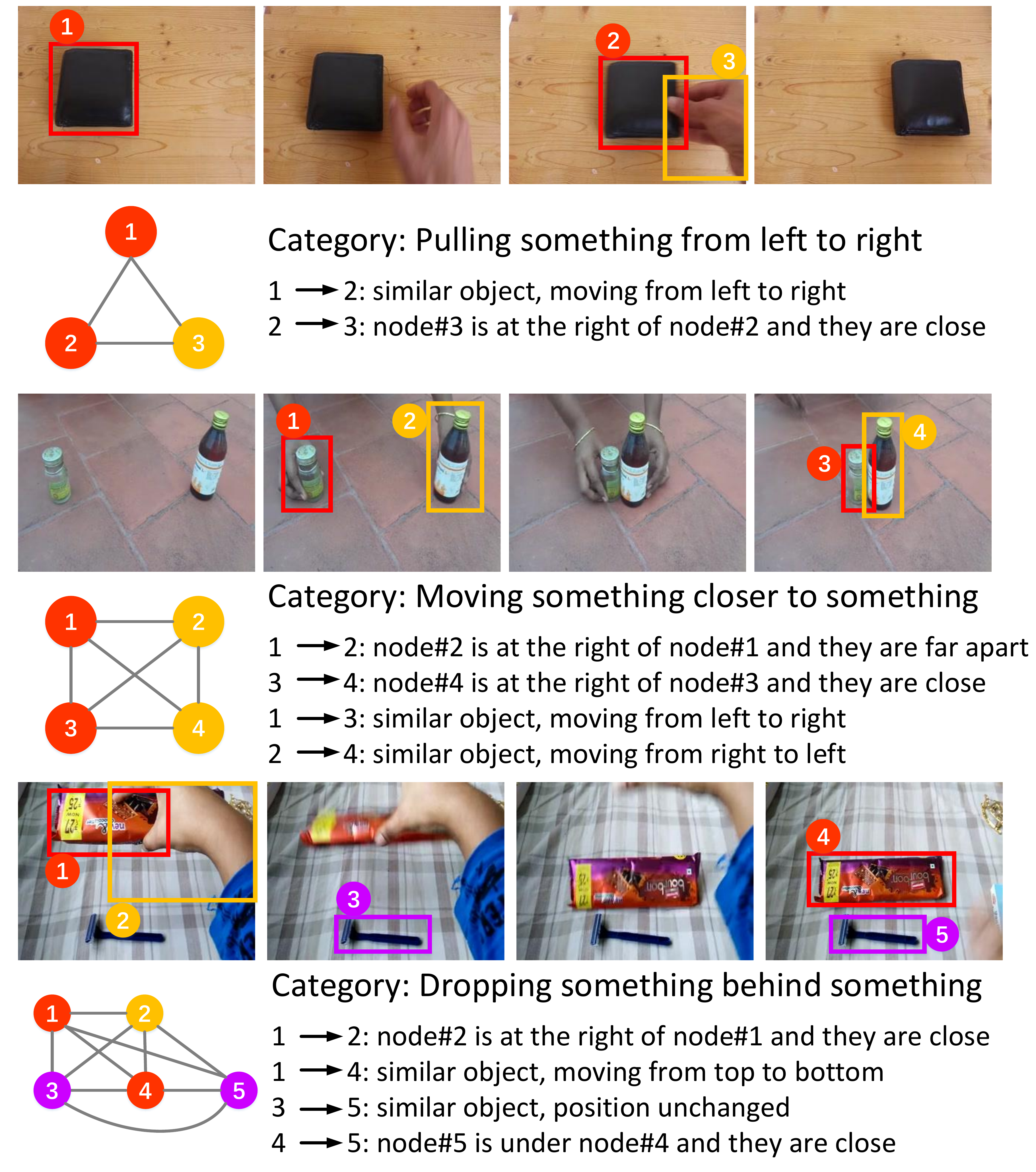}}
\vspace{-0.05in}
\caption{\small Examples showing the most similar sub-graphs with regard to the learnt action prototypes for three test videos. We only show the middle frame for each sampled video clip. Each sub-graph is equipped with several intuitive interpretations of semantic and relative spatial relations behind the graph edges.}
\label{fig:exp}
\vspace{-0.18in}
\end{figure}

Figure \ref{fig:exp} showcases the most similar sub-graphs w.r.t the learnt action prototypes in three test videos. We also display some intuitive interpretations of semantic and relative spatial relations behind the graph edges. From these exemplar results, it is easy to see that our learnt discriminative sub-graphs/action prototypes via MUSLE can capture the discriminative patterns of actions and thus facilitate action recognition. For instance, by encoding the patterns of ``two key objects are getting closer'' into the sub-graph of the second video, it is nature to recognize the action of ``Moving something closer to something''.

\textbf{Kinetics-400.}
To further verify the generality of our proposed MUSLE, we additionally conduct experiments for action recognition in the standard benchmark of Kinetics-400. Note that different from Something-Something which consists of object-centric videos with clean backgrounds, Kinetics-400 includes plenty of human-centric videos with complex backgrounds and some action categories (e.g., ``laughing'' and ``jogging'') are related to few objects. This naturally makes the reasoning over objects in Kinetics-400 more challenging than Something-Something. However, as shown in Table \ref{table:kinetics}, our MUSLE still manages to achieve competitive results against state-of-the-art techniques without graph-based reasoning. The results again verify the idea of exploiting multi-scale sub-graphs to reason about the dynamics of object interactions for action recognition.

\begin{table}[] \small
\centering
\caption{\small Comparison results of MUSLE with other state-of-the-art methods on Kinetics-400 validation set.}
\label{table:kinetics}
\begin{tabular}{l|c|cc}
\hline
\textbf{Method}               & \textbf{Backbone} & \textbf{Top-1} & \textbf{Top-5} \\ \hline\hline
TSN \cite{wang2016temporal}   & BNInception       & 69.1           & 88.7           \\
I3D \cite{carreira2017quo}    & InceptionV1       & 71.1           & 89.3           \\
R(2+1)D \cite{tran2018closer} & ResNet-34         & 72.0           & 90.0           \\
LGD \cite{qiu2019learning}    & ResNet-50         & 73.1           & 91.2           \\
ECO \cite{zolfaghari2018eco}  & BNInc+ResNet-18   & 70.7           & 89.4           \\
TSM \cite{lin2019tsm}         & ResNet-50         & 72.5           & 90.7           \\
STM \cite{jiang2019stm}       & ResNet-50         & 73.7           & 91.6           \\
TEA \cite{li2020tea}          & ResNet-50         & 74.0           & 91.3           \\ \hline
MUSLE                         & ResNet-50         & \textbf{75.1}  & \textbf{92.0}  \\ \hline
\end{tabular}
\end{table}

\begin{table}[] \small
\centering
\caption{\small Performance comparison on UCF101 and HMDB51.}
\label{table:cross}
\begin{tabular}{l|c|c|cc}
\hline
\textbf{Method} & \textbf{Pre-train} & \textbf{Finetune} & \textbf{UCF101} & \textbf{HMDB51} \\ \hline\hline
LGD \cite{qiu2019learning}            & Kinetics           & No                & 80.5            &  52.3               \\
LGD \cite{qiu2019learning}            & Kinetics           & Yes               & 93.2            &  70.1               \\ \hline
MUSLE           & Kinetics           & No                & 90.1            &  65.8               \\
MUSLE           & Kinetics           & Yes               & 94.8            &  72.2               \\ \hline
\end{tabular}
\vspace{-0.15in}
\end{table}

\subsection{Cross-Dataset Validation}
Recall that our MUSLE learns to discover prototypical sub-graphs/action prototypes that reflect the discriminative patterns for each human action. Such learnt action prototypes are generic, and could potentially be dataset-invariant, i.e., benefiting the recognition of actions in a target dataset different from the source training dataset. To evaluate this idea, we perform cross-dataset validation by pre-learning action prototypes on Kinetics-400 dataset and further leveraging them to classify videos in UCF101 \cite{soomro2012ucf101} and HMDB51 \cite{kuehne2011hmdb} datasets without or with fine-tuning. Here we select 44 and 16 action categories in UCF101 and HMDB51 respectively for evaluation, which also appear in the source training dataset (Kinetics-400). Details of the selected action categories in each target dataset can be referred in supplementary material. As shown in Table \ref{table:cross}, under the most challenging setting without any fine-tuning, the accuracy of our MUSLE can achieve 90.1\% and 65.8\% on UCF101 and HMDB51, making 9.6\% and 13.5\% absolute improvements over LGD. Moreover, when further fine-tuning on target datasets, our MUSLE still outperforms LGD. The results generally confirm the transferability of learnt action prototypes across different datasets in our MUSLE.

\section{Conclusion}
We have proposed MUlti-scale Sub-graph LEarning (MUSLE) framework, which explores the discriminative patterns of each action for recognition in videos. Particularly, we study the problem from the viewpoint of presenting the video structure as a space-time graph, decomposing such graph into sub-graphs, and interpreting the discriminative ones as action prototypes. To materialize our idea, we first link the region proposals across frames as tubelets, and build the spatio-temporal graph with nodes of tubelets and edges of dense connectivity. The graph is decomposed into multi-scale sub-graphs to characterize the complexity of different actions. Then, MUSLE learns a Gaussian Mixture Layer to estimate the distribution of sub-graphs from one certain action on each scale and takes Gaussian kernels as the prototypical structures of such action for recognition. Experiments conducted on three datasets, i.e., Something-Something V1 \& V2, and Kinetics-400, validate our proposal and analysis. More remarkably, our MUSLE achieves the superior accuracy on Something-Something V2.

\textbf{Acknowledgments.} This work was supported by the National Key R\&D Program of China under Grant No. 2020AAA0108600.

{\small
\bibliographystyle{ieee_fullname}
\bibliography{egbib}
}

\end{document}